\documentclass[11pt,twocolumn]{article}

\usepackage[
  top=1in,
  bottom=1in,
  left=0.75in,
  right=0.75in,
  columnsep=0.25in
]{geometry}

\usepackage{graphicx} 
\usepackage{amsmath}
\usepackage{amsfonts}
\usepackage{booktabs}
\usepackage{amsthm}
\usepackage{verbatim}
\usepackage{colortbl}
\usepackage{xcolor}
\usepackage{float}
\usepackage{natbib}
\usepackage{hyperref}

\newtheorem{remark}{Remark}

\title{Statistical benchmarking of transformer models in low signal-to-noise time-series forecasting}

\author{
Cyril Garcia \and Guillaume Remy
}

\date{\today}

\begin{document}

\maketitle

\begin{abstract}

We study the performance of transformer architectures for multivariate time-series forecasting in low-data regimes consisting of only a few years of daily observations. Using synthetically generated processes with known temporal and cross-sectional dependency structures and varying signal-to-noise ratios, we conduct bootstrapped experiments that enable direct evaluation via out-of-sample correlations with the optimal ground-truth predictor. We show that two-way attention transformers, which alternate between temporal and cross-sectional self-attention, can outperform standard baselines—Lasso, boosting methods, and fully connected multilayer perceptrons—across a wide range of settings, including low signal-to-noise regimes. We further introduce a dynamic sparsification procedure for attention matrices applied during training, and demonstrate that it becomes significantly effective in noisy environments, where the correlation between the target variable and the optimal predictor is on the order of a few percent. Analysis of the learned attention patterns reveals interpretable structure and suggests connections to sparsity-inducing regularization in classical regression, providing insight into why these models generalize effectively under noise.

\end{abstract}

\section{Introduction}

Time-series forecasting is the cornerstone of data-driven decision-making in multiple domains, from finance and meteorology to supply chain and energy management. Traditional statistical models, such as ARIMA and exponential smoothing \cite{box2015time}, have long dominated this field due to their interpretability and efficiency on univariate data. However, with the explosion of multivariate, high-dimensional datasets, these methods increasingly fall short in the presence of complex interdependencies across time and features.
In recent years, transformer architectures—which originally revolutionized natural language processing through self-attention mechanisms \cite{vaswani2017attention}—have shown promise in time-series tasks and have led to numerous architecture proposals for forecasting, as reviewed in \cite{wen2023transformers}. Notable examples include the Informer model for efficient long-sequence forecasting \cite{zhou2021informer}, Autoformer with decomposition and auto-correlation mechanisms \cite{wu2021autoformer}, FEDformer leveraging frequency-enhanced attention \cite{zhou2022fedformer}, and PatchTST for patch-based representations \cite{nie2023time}. Yet, recently, the effectiveness of transformers for time-series forecasting has been questioned, with works showing that simpler (linear) models can achieve comparable or superior performance \cite{zeng2023transformers,das2023tide} on many real-world benchmark datasets.

To better understand the conditions under which transformer-based models actually outperform alternatives, we propose a controlled benchmarking framework using synthetic data, where both noise levels and types of dependencies in the forecasting problem can be varied, and the performance of a range of model choices can be statistically evaluated. More precisely, we frame the forecasting problem as predicting a target series $Y_{t,n}$ from predictors $X^{(j)}_{s,n}$, incorporating dimensions for time indexed by $s,t$ (with $s \leq t$, ensuring causality), series indexed by $n$, and features indexed by $j$. To dissect model behaviors, we generate artificial datasets with tunable ``orders'' of effects: \textit{Order 0} (simple linear dependencies), \textit{Order 1} (shifts in time or cross-section, or non-linear feature interactions), and \textit{Order 2} (combined shifts, non-linear interactions in the time-series or cross-sectional dimension). Noise is systematically varied, and we measure model performance using out-of-sample correlations against the optimal ground-truth dependency. A key assumption of this paper is that the time series of predictors $X$ is stationary and has already been preprocessed and normalized. This allows us to focus solely on evaluating the transformers' ability to capture dependencies at different noise levels, leaving the pre-processing stage for a later study.

Our contributions are twofold. First, we test several transformer architectures with two-way attention mechanisms—two-way referring to applying attention along both the time-series and cross-sectional dimensions (as in \cite{ho2019axial, liu2024itransformer})—to exploit the multi-dimensional structure of time-series data. We benchmark these models against baselines like Lasso regression \cite{tibshirani1996regression}, boosting \cite{friedman2001greedy}, as well as simpler neural network architectures, revealing that transformers outperform traditional methods on a number of dependencies and are general enough to work in several instances of data generating processes, even at high noise levels. Second, we propose and test a sparse attention implementation \cite{child2019generating} to enhance robustness in low-signal regimes, which yields up to 10--20$\%$ performance gains in correlation with the optimal predictor. Additionally, we provide in Appendix \ref{app_theo} an analytical computation that gives the expected correlation in the linear case, bridging empirical results with statistical theory.

The remainder of the paper is organized as follows. Section 2 details the problem setup, data generation, model evaluation, and transformer architectures. Section 3 presents the experimental results when varying noise levels and time-series dependencies. Section 4 introduces our dynamic sparse attention implementation and presents the associated experimental results. Lastly, Section \ref{sec_conclu} gives a conclusion and future outlooks.

\section{Setup}\label{sec_setup}

\subsection{The forecasting problem}

We start by considering the following general time-series forecasting problem of the target series $Y$ using the time-series of predictors $X$:
\begin{align}\label{forecasting_problem}
Y_{t,n} = \mathcal{F} \left( \left( X_{s,n}^{(j)} \right)_{s \leq t, \; n \leq N, \; j \leq F} \right) + \epsilon.
\end{align}
Here the indices $s,t$ and $n$ denote respectively the time and series indices, they obey $1 \leq s \leq t \leq T$ and $1 \leq n \leq N$. We will refer to them frequently as the time-series and cross-sectional directions. The series $X$ also contains a features dimension indexed by $j$ satisfying $1 \leq j \leq F$. The quantity $Y_{t,n}$ can depend a priori on an arbitrary function $\mathcal{F}$ of all the $X_{s,n}^{(j)}$ plus some noise $\epsilon$, the only constraint being $s \leq t$ implying the temporal dependence can only be up to the present time $t$. In this paper we also assume stationarity of the series, namely that the functional dependence $\mathcal{F}$ is invariant by shifts $t \rightarrow t + t_0$, $t_0 \geq 0$ (see below the concrete examples of $\mathcal{F}$).

\begin{remark}
To give an example of what the $T,N,F$ dimensions could correspond to, assume a forecasting problem where one is interested in predicting the temperature in different cities on different days. For this it is natural to set:
\begin{itemize}
\item \textbf{T}: The time index $t$ indexes the days of the year.
\item \textbf{N}: The series index $n$ corresponds to the different cities or locations.
\item \textbf{F}: The feature index $j$ corresponds to the different available features for forecasting, which could be for instance temperature, wind, humidity, etc...
\end{itemize}
$X_{t,n}^{(j)}$ then contains all the data above and $Y_{t,n}$ would correspond to the one day forward temperature we wish to predict. In this setup if $j=1$ corresponds to the temperature feature one would have $Y_{t,n} = X_{t+1, n}^{(1)}$ but in general this doesn't need to be the case.
In some setups it is also possible to have a single feature, i.e. $F=1$, or that the target time-series $Y$ is only one-dimensional, i.e. $N=1$, but we keep all three dimensions to treat the most general case.
\end{remark}

\subsection{Types of time-series dependencies}

To make progress in understanding this forecasting problem we now make some assumptions on the function $\mathcal{F}$. We first rewrite \eqref{forecasting_problem} as $Y_{t,n} = \widetilde Y_{t,n} + \epsilon$, where $\widetilde Y_{t,n}$ is the ground-truth or optimal prediction of the model, and assume  a correlation of $\rho = \text{Correl}(Y_{t,n}, \widetilde Y_{t,n})$. This $\rho$ is the key parameter that controls the signal-to-noise ratio in the experiments that we will run. We then further assume $\widetilde Y_{t,n}$ can be written as a sum of simple effects indexed by $e$:
\begin{align}\label{eq_sum_effects}
\widetilde Y_{t,n} = \sum_e \rho_e \widetilde Y_{t,n}^{e}.
\end{align}
Here $\rho_e \in [0,1], \rho^2 = \sum_e \rho_e^2,$ and $e$ will belong to a list of typical effects that are expected in a time-series forecasting problem which we now describe. The simplest effect $e = \text{Lin}$, which we will call the \textit{Order 0}, corresponds to assuming the linear dependence
\begin{align}\label{lin_effect}
 \widetilde Y_{t,n}^{\text{Lin}} = \sum_j \rho_{j,n} X_{t,n}^{(j)},
\end{align}
for $\rho_{j,n} \in [0,1]$ and for every $t,n$. In practice $\rho_{j, n}$ could be independent of $n$ if the effect is assumed to be the same for all series. Beyond this linear case we can consider shifted relationships, either in the time-series (TS) or cross-section (CS) directions, namely
\begin{align}\label{TS_shift}
\widetilde Y_{t,n}^{\text{TS-Shift}} = \sum_j \rho_{j,n} X_{t - s_{j,n},n}^{(j)},
\end{align}
\begin{align}\label{CS_shift}
\widetilde Y_{t,n}^{\text{CS-Shift}} = \sum_j \rho_{j,n} X_{t,n+k_{j,n}}^{(j)},
\end{align}
where in the first case the shifts $s_{j,n} \geq 0$ need to be non-negative, and in the second case the sum $n + k_{j,n}$ should be understood modulo the number $N$ of series. Alternatively we can also replace the linear sum of \eqref{lin_effect} by a function $G: \mathbb{R} ^F \rightarrow \mathbb{R}$ of the features, for $\rho_{\text{Fea-Nonlin}} \in [0,1]$:
\begin{align}\label{Non-lin}
\widetilde Y_{t,n}^{\text{Fea-Nonlin}} = \rho_{\text{Fea-Nonlin}} \; G(X_{t,n}^{(1)}, \dots, X_{t,n}^{(F)}).
\end{align}
As a very basic example of $G$ we could take $X_{t,n}^{(j_1)} \textrm{sign} (X_{t,n}^{(j_2)}) $, the conditioning of feature $j_1$ by the sign of feature $j_2$. Another simple case is any non-linear function of just one of the features, i.e. $G(X_{t,n}^{(j_1)})$.
We call (\ref{TS_shift}), (\ref{CS_shift}), and (\ref{Non-lin}) the \textit{Order 1} effects. Lastly we consider the following three effects
\begin{align}\label{tscs_shift}
&\widetilde Y_{t,n}^{\text{TSCS-Shift}} = \sum_j \rho_{j,n} X_{t-s_{j,n},n+k_{j,n}}^{(j)},\\
&\widetilde Y_{t,n}^{\text{TS-Nonlin}} = \sum_j \rho_{j,n} G(X_{t,n}^{(j)}, \dots, X_{t-s,n}^{(j)}),\\
&\widetilde Y_{t,n}^{\text{CS-Nonlin}} = \sum_j \rho_{j,n} G(X_{t,1}^{(j)}, \dots, X_{t,N}^{(j)}),
\end{align}
which correspond respectively to shifting simultaneously in the time-series and cross-sectional directions, considering a non-linear interaction $G$ between time-series lags up to lag $t-s$, or a non-linear interaction $G$ in the cross-sectional direction. We classify these three effects as \textit{Order 2}. It can be argued that since through equation \eqref{eq_sum_effects} we are taking an arbitrary sum of all the effects described above, a very general class of dependencies $\mathcal{F}$ can be approximated.

\subsection{Generating artificial data and model evaluation}

Based on the hypotheses made above on $X,Y$ it is straightforward to generate this data with the following steps.
\begin{itemize}
\item Sample $X_{t,n}^{(j)}$ i.i.d. $\mathcal{N}(0,1)$ for all indices $t,n,j$.
\item Following \eqref{eq_sum_effects}, combine the $X_{t,n}^{(j)}$ using one or a sum of effects presented above and obtain $\widetilde Y_{t,n}$.
\item Assuming $\widetilde Y_{t,n}$ has been constructed to have mean $0$ and variance $\rho^2 < 1$, define $Y_{t,n} = \widetilde Y_{t,n} + \sqrt{1 - \rho^2} Z_{t,n}$, with $Z_{t,n}$ i.i.d. $\mathcal{N}(0,1)$. 
\end{itemize}
This procedure gives an optimal prediction $\widetilde Y_{t,n}$ of mean $0$ and variance $\rho^2$, and a target $Y_{t,n}$ of mean $0$, variance $1$, and correlation $\rho$ with $\widetilde Y_{t,n}$. A model will then, given $X_{s,n}^{(j)}$ for all $n,j$ and $s \leq t$, output $\widehat Y_{t,n} $, a prediction for $Y_{t, n}$. We split the data points into train and test, $T = T_{\text{train}} + T_{\text{test}}$, train the model using the $T_{\text{train}}$ points, and evaluate the model by computing the correlation between $\widehat Y_{t,n}$ and the optimal prediction $\widetilde Y_{t,n}$ using the $T_{\text{test}}$ points.

To fit our models we introduce a maximum lookback window length $T_{\text{win}}$, namely to make a prediction at time step $t$ the model will only use the $T_{\text{win}}$ time steps $t-T_{\text{win}} +1 \leq s \leq t$. We also fix the dimensions of the data. Unless specified otherwise we use $T_{\text{train}}=2500$, $T_{\text{test}}=1500$, $N=10$, $F=20$, $T_{\text{win}} =10$. Note that these dimensions are chosen such that a dataset containing daily observations over roughly ten years would be the same order of magnitude as the data studied here which should cover a rather wide range of practical examples. This is also constraining as this is a relatively small amount of data when compared to the usual experimental datasets studied (see \cite{wu2021autoformer, zhou2021informer} for references to many such datasets).

\subsection{OLS, Lasso, Boosting, MLP as forecasting baseline}\label{sec_baselines}

Before describing the different transformer architectures, we give a list of simple linear and non-linear models we can fit to obtain a simple performance baseline. In these benchmark models, since there is no explicit way to consider the time or cross-sectional dimensions, we fit those benchmarks by flattening the three dimensions $T_{\text{win}}, N, F$. This greedy approach considers a relatively large number of features and can be expected to be limited by the curse of dimensionality as dimensions increase, but this gives nevertheless a good baseline.

\textbf{i) Ordinary least squares (OLS).}
First we can fit an OLS model ignoring the time-series structure.
In this case, assuming only linear effects are present, one can theoretically compute the expected model performance. Let $\gamma$ denote the total number of features divided by the number of time points we have for prediction (fitting here one OLS model per time series to predict):
\begin{align}
\gamma = \frac{T_{\text{win}} \cdot N \cdot F}{T_{\text{train}}}.
\end{align}
In Appendix \ref{app_theo} it is derived that the expected out-of-sample correlation to $\widetilde Y_{t,n}$ for this OLS prediction is given by:
\begin{align}\label{eq_theo}
C(\rho, \gamma) = \frac{\rho}{\sqrt{\rho^2 + (1- \rho^2) \frac{\gamma}{1-\gamma}}}.
\end{align}
We call this quantity TheoC in the result tables, it gives the expected amount of correlation when fitting an OLS per time-series using all the features.

\textbf{ii) Global Lasso regression / Boosting.}
Next, we can run a global Lasso regression  where the Lasso penalty parameter is estimated using a standard cross-validation. Here global refers to the fact that we treat all time-series as examples (instead of fitting one model per target time-series as in \textbf{i)}). More precisely, we use a unique set of regression betas $\beta_{s,n',j}$ indexed by lags, series, features to make all $N$ time-series prediction. As a simple non-linear ML benchmark we will also fit a standard boosting model using again all features, similarly as for the global Lasso. This gives a standard benchmark which has the possibility of capturing non-linear effects.

\textbf{iii) Global multi-layer perceptron.}
Lastly, we move to a multi-layer perceptron (MLP), assuming no time-series structure, which we call the global MLP.
This model flattens the input $X_{s,n}^{(j)} \in \mathbb{R}^{T_{\text{win}} \times N \times F}$ into a vector of dimension $T_{\text{win}} \times N \times F$ and applies four 512-dimensional linear layers with GELU activations and $0.1$ dropout, followed by a final projection to $N$ outputs. This model gives a simple neural network baseline without taking advantage of the time-series structure.

\subsection{Transformer architectures for forecasting}\label{subsec_models}

We now propose a general transformer architecture for time-series forecasting, a model which applies self-attention along the time-series and cross-sectional dimensions following a custom ordering of time-series and cross-sectional attention blocks. Given an  input $\mathbf{X} \in \mathbb{R}^{B \times T_{\text{win}} \times N \times F}$ of dimensions batch size, lookback window length, number of time-series, number of features, the architecture proceeds as follows:

\begin{enumerate}
    \item The feature dimension $F$ is linearly projected to $d_{\text{model}} = 64$.
    \item Two learned positional embeddings are added: one for time steps and one for the $N$ series.
    \item The core consists of $L$ attention blocks defined by a string (e.g., \texttt{"TCTC"}):
    \begin{itemize}
        \item \textbf{T-block}: Independent temporal attention over the $T_{\text{win}}$ dimension for each of the $N$ series (shape: $B \cdot N \times T_{\text{win}} \times d_{\text{model}}$).
        \item \textbf{C-block}: Independent cross-sectional attention over the $N$ variables at each time step (shape: $B \cdot T_{\text{win}} \times N \times d_{\text{model}}$).
    \end{itemize}
    Each attention uses $8$ attention heads, and feeds into a feed-forward block with internal dimension 256.
    \item An output head (LayerNorm $\to$ GELU $\to$ Linear) produces the final prediction for all $N$ targets.
\end{enumerate}
A dropout of 0.1 is used for training. In the results table we call this model Trans, and unless stated otherwise it is run in the configuration \texttt{"TCTC"}, namely stacking twice a T-block followed by a C-block.

\section{Experiments}\label{sec_results}

We now present the results of the experiments. In all the tables below $\rho$ always corresponds to the global signal-to-noise ratio, namely the correlation between $\widetilde Y_{t,n}$ and $Y_{t,n}$. The rest of this section is split into two subsections, based on either fitting the models on a $Y_{t,n}$ generated from a single effect or on all the effects simultaneously.

\subsection{Model performance on each individual effect}

Here we construct our data $X, Y$ by choosing: $T_{\text{train}} = 2500$, $T_{\text{test}} = 1500$, $T_{\text{win}} = 10$, $N = 10$, $F = 20$. We pick as values of $\rho$: $2\%, 5\%, 10\%, 20\%, 50\%$. When building $\widetilde Y_{t,n}$ we only use a single effect, namely the sum \eqref{eq_sum_effects} is restricted to a single effect $e$ in the list Lin, TS-Shift, CS-Shift, Fea-Nonlin, TSCS-Shift. Furthermore we only use half of the features when building $\widetilde Y_{t,n}$ from $X$, leaving the other half as noise. We give one result table for each effect and each entry in the tables corresponds to the out-of-sample performance of a model, which is measured as the correlation between the model prediction $\widehat Y_{t,n}$ and the ground-truth optimal prediction $\widetilde Y_{t,n}$, computed over the $T_{\text{test}}$ data points. Each result table contains five models as rows: TheoC corresponding to the value given by \eqref{eq_theo}, Lasso, Boosting and MLP corresponding to the models presented in Section \ref{sec_baselines}, and lastly our transformer model Trans described in Section \ref{subsec_models}. We now comment effect by effect on the results obtained:
\begin{itemize}
    \item \textbf{Linear effect.} Generated from equation \eqref{lin_effect}. Table \ref{table_lin} shows that Lasso regression is the best model in this case which is expected since we are dealing with a linear relationship, but the transformer model performs quite well at small $\rho$, especially compared to the Boosting and MLP.
    \item \textbf{TS-Shift effect.} Generated from equation \eqref{TS_shift} with the lag $s_{j,n}=1$ for simplicity. Table \ref{table_ts_shift} shows Lasso performs well here as in the linear case for the same reason, but the transformer model Trans does not perform quite as well on TS-Shift as on the Linear effect.
    \item \textbf{CS-Shift effect.} Generated from equation \eqref{CS_shift} where the cross-sectional shift $k_{j,n}$ is chosen at random for each $j,n$, namely each feature of each time-series predicts another time-series chosen among all possible. Table \ref{table_cs_shift} shows that the transformer model has a great performance here, the cross-sectional attention detects this shift very well comparatively to the benchmarks.
    \item \textbf{Fea-Nonlin effect.} Generated from equation \eqref{Non-lin} with $G=X_{t,n}^{(j_1)} \textrm{sign} (X_{t,n}^{(j_2)})$, namely the conditioning of one feature by the sign of another. This is a non-linear effect. Table \ref{table_nonlin} shows that Lasso does not work here (expected), while the transformer performs very well. Both Boosting and MLP fail here because they flatten all features and the dimensionality is then too big to find the non-linear effect.
    \item \textbf{TSCS-Shift effect.}  Generated from equation \eqref{tscs_shift} using the same rules as above for both the TS and CS shifts, namely set $s_{j,n}=1$ and choose $k_{j,n}$ at random. Table \ref{table_tscs_shift} shows that here interestingly the MLP is the best model, since it assumes no time-series structure it obtains similar results on the TS, CS, and TSCS shifts. On the other hand the transformer architecture performs very poorly here.
\end{itemize}

\begin{table}[t]
    \centering
    \begin{tabular}{lccccc}
\toprule
$\rho$ & 0.02 & 0.05 & 0.10 & 0.20 & 0.50 \\
\midrule
TheoC & {\cellcolor[HTML]{3B4CC0}} \color[HTML]{F1F1F1} 0.010 & {\cellcolor[HTML]{3B4CC0}} \color[HTML]{F1F1F1} 0.025 & {\cellcolor[HTML]{3B4CC0}} \color[HTML]{F1F1F1} 0.050 & {\cellcolor[HTML]{3B4CC0}} \color[HTML]{F1F1F1} 0.102 & {\cellcolor[HTML]{3B4CC0}} \color[HTML]{F1F1F1} 0.277 \\
Lasso & {\cellcolor[HTML]{EE8468}} \color[HTML]{F1F1F1} 0.038 & {\cellcolor[HTML]{B40426}} \color[HTML]{F1F1F1} 0.305 & {\cellcolor[HTML]{B40426}} \color[HTML]{F1F1F1} 0.785 & {\cellcolor[HTML]{B40426}} \color[HTML]{F1F1F1} 0.940 & {\cellcolor[HTML]{B40426}} \color[HTML]{F1F1F1} 0.995 \\
Boosting & {\cellcolor[HTML]{4358CB}} \color[HTML]{F1F1F1} 0.011 & {\cellcolor[HTML]{5E7DE7}} \color[HTML]{F1F1F1} 0.057 & {\cellcolor[HTML]{6F92F3}} \color[HTML]{F1F1F1} 0.173 & {\cellcolor[HTML]{DDDCDC}} \color[HTML]{000000} 0.521 & {\cellcolor[HTML]{E26952}} \color[HTML]{F1F1F1} 0.892 \\
MLP & {\cellcolor[HTML]{ADC9FD}} \color[HTML]{000000} 0.022 & {\cellcolor[HTML]{536EDD}} \color[HTML]{F1F1F1} 0.047 & {\cellcolor[HTML]{506BDA}} \color[HTML]{F1F1F1} 0.104 & {\cellcolor[HTML]{506BDA}} \color[HTML]{F1F1F1} 0.162 & {\cellcolor[HTML]{82A6FB}} \color[HTML]{F1F1F1} 0.435 \\
Trans & {\cellcolor[HTML]{B40426}} \color[HTML]{F1F1F1} 0.045 & {\cellcolor[HTML]{C5D6F2}} \color[HTML]{000000} 0.141 & {\cellcolor[HTML]{82A6FB}} \color[HTML]{F1F1F1} 0.212 & {\cellcolor[HTML]{97B8FF}} \color[HTML]{000000} 0.336 & {\cellcolor[HTML]{EFCFBF}} \color[HTML]{000000} 0.693 \\
\bottomrule
\end{tabular}
\vskip 0.05in
\caption{Model performance for effect Linear}
    \label{table_lin}
    \end{table}

\begin{table}[t]\label{table_ts_shift}
    \centering
    \begin{tabular}{lccccc}
\toprule
$\rho$ & 0.02 & 0.05 & 0.10 & 0.20 & 0.50 \\
\midrule
TheoC & {\cellcolor[HTML]{4F69D9}} \color[HTML]{F1F1F1} 0.010 & {\cellcolor[HTML]{3B4CC0}} \color[HTML]{F1F1F1} 0.025 & {\cellcolor[HTML]{3B4CC0}} \color[HTML]{F1F1F1} 0.050 & {\cellcolor[HTML]{3B4CC0}} \color[HTML]{F1F1F1} 0.102 & {\cellcolor[HTML]{3B4CC0}} \color[HTML]{F1F1F1} 0.277 \\
Lasso & {\cellcolor[HTML]{B40426}} \color[HTML]{F1F1F1} 0.038 & {\cellcolor[HTML]{B40426}} \color[HTML]{F1F1F1} 0.315 & {\cellcolor[HTML]{B40426}} \color[HTML]{F1F1F1} 0.812 & {\cellcolor[HTML]{B40426}} \color[HTML]{F1F1F1} 0.960 & {\cellcolor[HTML]{B40426}} \color[HTML]{F1F1F1} 0.996 \\
Boosting & {\cellcolor[HTML]{DDDCDC}} \color[HTML]{000000} 0.023 & {\cellcolor[HTML]{6A8BEF}} \color[HTML]{F1F1F1} 0.069 & {\cellcolor[HTML]{86A9FC}} \color[HTML]{F1F1F1} 0.227 & {\cellcolor[HTML]{EAD5C9}} \color[HTML]{000000} 0.576 & {\cellcolor[HTML]{E26952}} \color[HTML]{F1F1F1} 0.893 \\
MLP & {\cellcolor[HTML]{3B4CC0}} \color[HTML]{F1F1F1} 0.008 & {\cellcolor[HTML]{485FD1}} \color[HTML]{F1F1F1} 0.038 & {\cellcolor[HTML]{4B64D5}} \color[HTML]{F1F1F1} 0.093 & {\cellcolor[HTML]{5875E1}} \color[HTML]{F1F1F1} 0.183 & {\cellcolor[HTML]{84A7FC}} \color[HTML]{F1F1F1} 0.438 \\
Trans & {\cellcolor[HTML]{445ACC}} \color[HTML]{F1F1F1} 0.009 & {\cellcolor[HTML]{80A3FA}} \color[HTML]{F1F1F1} 0.087 & {\cellcolor[HTML]{7295F4}} \color[HTML]{F1F1F1} 0.181 & {\cellcolor[HTML]{93B5FE}} \color[HTML]{000000} 0.331 & {\cellcolor[HTML]{F5C1A9}} \color[HTML]{000000} 0.734 \\
\bottomrule
\end{tabular}
\vskip 0.05in
\caption{Model performance for effect TS-Shift}
    \label{table_ts_shift}
    \end{table}

\begin{table}[t]\label{table_cs_shift}
    \centering
    \begin{tabular}{lccccc}
\toprule
$\rho$ & 0.02 & 0.05 & 0.10 & 0.20 & 0.50 \\
\midrule
TheoC & {\cellcolor[HTML]{E8D6CC}} \color[HTML]{000000} 0.010 & {\cellcolor[HTML]{6F92F3}} \color[HTML]{F1F1F1} 0.025 & {\cellcolor[HTML]{98B9FF}} \color[HTML]{000000} 0.050 & {\cellcolor[HTML]{7597F6}} \color[HTML]{F1F1F1} 0.102 & {\cellcolor[HTML]{7093F3}} \color[HTML]{F1F1F1} 0.277 \\
Lasso & {\cellcolor[HTML]{B5CDFA}} \color[HTML]{000000} 0.006 & {\cellcolor[HTML]{D65244}} \color[HTML]{F1F1F1} 0.047 & {\cellcolor[HTML]{EE8468}} \color[HTML]{F1F1F1} 0.099 & {\cellcolor[HTML]{B40426}} \color[HTML]{F1F1F1} 0.288 & {\cellcolor[HTML]{F0CDBB}} \color[HTML]{000000} 0.435 \\
Boosting & {\cellcolor[HTML]{3B4CC0}} \color[HTML]{F1F1F1} -0.002 & {\cellcolor[HTML]{3B4CC0}} \color[HTML]{F1F1F1} 0.020 & {\cellcolor[HTML]{3B4CC0}} \color[HTML]{F1F1F1} 0.023 & {\cellcolor[HTML]{3B4CC0}} \color[HTML]{F1F1F1} 0.061 & {\cellcolor[HTML]{3B4CC0}} \color[HTML]{F1F1F1} 0.214 \\
MLP & {\cellcolor[HTML]{B40426}} \color[HTML]{F1F1F1} 0.020 & {\cellcolor[HTML]{EE8468}} \color[HTML]{F1F1F1} 0.044 & {\cellcolor[HTML]{F7AF91}} \color[HTML]{000000} 0.089 & {\cellcolor[HTML]{ECD3C5}} \color[HTML]{000000} 0.189 & {\cellcolor[HTML]{E8D6CC}} \color[HTML]{000000} 0.419 \\
Trans & {\cellcolor[HTML]{D44E41}} \color[HTML]{F1F1F1} 0.018 & {\cellcolor[HTML]{B40426}} \color[HTML]{F1F1F1} 0.050 & {\cellcolor[HTML]{B40426}} \color[HTML]{F1F1F1} 0.118 & {\cellcolor[HTML]{F7AF91}} \color[HTML]{000000} 0.218 & {\cellcolor[HTML]{B40426}} \color[HTML]{F1F1F1} 0.589 \\
\bottomrule
\end{tabular}
\vskip 0.05in
\caption{Model performance for effect CS-Shift}
    \label{table_cs_shift}
    \end{table}

\begin{table}[t]\label{table_nonlin}
    \centering
    \begin{tabular}{lccccc}
\toprule
$\rho$ & 0.02 & 0.05 & 0.10 & 0.20 & 0.50 \\
\midrule
TheoC & {\cellcolor[HTML]{AAC7FD}} \color[HTML]{000000} 0.010 & {\cellcolor[HTML]{EDD1C2}} \color[HTML]{000000} 0.025 & {\cellcolor[HTML]{CAD8EF}} \color[HTML]{000000} 0.050 & {\cellcolor[HTML]{D6DCE4}} \color[HTML]{000000} 0.102 & {\cellcolor[HTML]{EDD2C3}} \color[HTML]{000000} 0.277 \\
Lasso & {\cellcolor[HTML]{3B4CC0}} \color[HTML]{F1F1F1} -0.001 & {\cellcolor[HTML]{3B4CC0}} \color[HTML]{F1F1F1} -0.010 & {\cellcolor[HTML]{4A63D3}} \color[HTML]{F1F1F1} 0.001 & {\cellcolor[HTML]{5673E0}} \color[HTML]{F1F1F1} 0.005 & {\cellcolor[HTML]{3F53C6}} \color[HTML]{F1F1F1} 0.001 \\
Boosting & {\cellcolor[HTML]{94B6FF}} \color[HTML]{000000} 0.008 & {\cellcolor[HTML]{688AEF}} \color[HTML]{F1F1F1} -0.001 & {\cellcolor[HTML]{4055C8}} \color[HTML]{F1F1F1} -0.003 & {\cellcolor[HTML]{4F69D9}} \color[HTML]{F1F1F1} -0.001 & {\cellcolor[HTML]{7396F5}} \color[HTML]{F1F1F1} 0.082 \\
MLP & {\cellcolor[HTML]{6A8BEF}} \color[HTML]{F1F1F1} 0.004 & {\cellcolor[HTML]{536EDD}} \color[HTML]{F1F1F1} -0.005 & {\cellcolor[HTML]{3B4CC0}} \color[HTML]{F1F1F1} -0.006 & {\cellcolor[HTML]{3B4CC0}} \color[HTML]{F1F1F1} -0.018 & {\cellcolor[HTML]{3B4CC0}} \color[HTML]{F1F1F1} -0.007 \\
Trans & {\cellcolor[HTML]{B40426}} \color[HTML]{F1F1F1} 0.032 & {\cellcolor[HTML]{B40426}} \color[HTML]{F1F1F1} 0.051 & {\cellcolor[HTML]{B40426}} \color[HTML]{F1F1F1} 0.124 & {\cellcolor[HTML]{B40426}} \color[HTML]{F1F1F1} 0.235 & {\cellcolor[HTML]{B40426}} \color[HTML]{F1F1F1} 0.493 \\
\bottomrule
\end{tabular}
\vskip 0.05in
\caption{Model performance for effect Fea-Nonlin}
    \label{table_nonlin}
    \end{table}

\begin{table}[t]\label{table_tscs_shift}
    \centering
    \begin{tabular}{lccccc}
\toprule
$\rho$ & 0.02 & 0.05 & 0.10 & 0.20 & 0.50 \\
\midrule
TheoC & {\cellcolor[HTML]{6485EC}} \color[HTML]{F1F1F1} 0.010 & {\cellcolor[HTML]{F5A081}} \color[HTML]{000000} 0.025 & {\cellcolor[HTML]{E6D7CF}} \color[HTML]{000000} 0.050 & {\cellcolor[HTML]{E4D9D2}} \color[HTML]{000000} 0.102 & {\cellcolor[HTML]{EED0C0}} \color[HTML]{000000} 0.277 \\
Lasso & {\cellcolor[HTML]{AAC7FD}} \color[HTML]{000000} 0.013 & {\cellcolor[HTML]{CDD9EC}} \color[HTML]{000000} 0.012 & {\cellcolor[HTML]{B40426}} \color[HTML]{F1F1F1} 0.092 & {\cellcolor[HTML]{B40426}} \color[HTML]{F1F1F1} 0.188 & {\cellcolor[HTML]{B40426}} \color[HTML]{F1F1F1} 0.460 \\
Boosting & {\cellcolor[HTML]{3B4CC0}} \color[HTML]{F1F1F1} 0.008 & {\cellcolor[HTML]{9ABBFF}} \color[HTML]{000000} 0.005 & {\cellcolor[HTML]{6788EE}} \color[HTML]{F1F1F1} 0.014 & {\cellcolor[HTML]{89ACFD}} \color[HTML]{000000} 0.049 & {\cellcolor[HTML]{CEDAEB}} \color[HTML]{000000} 0.221 \\
MLP & {\cellcolor[HTML]{B40426}} \color[HTML]{F1F1F1} 0.023 & {\cellcolor[HTML]{B40426}} \color[HTML]{F1F1F1} 0.037 & {\cellcolor[HTML]{C0282F}} \color[HTML]{F1F1F1} 0.089 & {\cellcolor[HTML]{C0282F}} \color[HTML]{F1F1F1} 0.182 & {\cellcolor[HTML]{CB3E38}} \color[HTML]{F1F1F1} 0.433 \\
Trans & {\cellcolor[HTML]{4F69D9}} \color[HTML]{F1F1F1} 0.009 & {\cellcolor[HTML]{3B4CC0}} \color[HTML]{F1F1F1} -0.008 & {\cellcolor[HTML]{3B4CC0}} \color[HTML]{F1F1F1} 0.001 & {\cellcolor[HTML]{3B4CC0}} \color[HTML]{F1F1F1} 0.005 & {\cellcolor[HTML]{3B4CC0}} \color[HTML]{F1F1F1} 0.029 \\
\bottomrule
\end{tabular}
\vskip 0.05in
\caption{Model performance for effect TSCS-Shift}
    \label{table_tscs_shift}
    \end{table}

Overall the transformer model strikes a good balance between capturing decently well linear effects and the TS-Shift, while being the only model detecting the non-linear feature interactions and having on average the best performance on the CS-shift. The only shortcoming is that it was not able to detect the TSCS-Shift effect. We leave understanding this problem for a future work.

\subsection{Model performance on sum of all effects}

Next we test the performance of our models at predicting a target $Y_{t,n}$ which simultaneously contains a superposition of 5 effects. Fix again $T_{\text{train}} = 2500$, $T_{\text{test}} = 1500$, $T_{\text{win}} = 10$, $N = 10$, $F = 20$. Now out of the 20 features, as before half of them will not be used to generate $\widetilde Y_{t,n}$, and out of the remaining 10 we attribute 2 to each of the five effects Lin, TS-Shift, CS-
Shift, Fea-Nonlin, TSCS-Shift. Hence our $\widetilde Y_{t,n}$ contains the superposition of the 5 effects, but with a correlation to $Y_{t,n}$ still fixed to $\rho$. Now when testing the model performance, on top of giving as before the correlation between the model prediction $\widehat Y_{t,n}$ and $\widetilde Y_{t,n}$ (``Optimal" column in the result tables), we can also take a look at the correlation between $\widehat Y_{t,n}$ and each $\widetilde Y_{t,n}^e$ for each effect $e$ in the list of effects (column with the corresponding effect name in the result tables). This allows us to analyze which effect a model focuses on when all are present at once. For these results we set $\rho$ equal to $0.05$ or $0.2$ and obtain Table \ref{tab:model_performance1} and Table \ref{tab:model_performance2}. These two tables show that the transformer model while not having the best correlation to the optimal prediction - Lasso is the best since it detects Linear and TS-Shift so well - shows a good performance simultaneously across all the effects at once, apart from the TSCS-shift it struggles to find.

\begin{table*}[t]
    \centering
    \begin{tabular}{lcccccc}
\toprule
Effect & Optimal & Linear & Fea-Nonlin & TS-Shift & CS-Shift & TSCS-Shift \\
\midrule
Lasso & {\cellcolor[HTML]{B40426}} \color[HTML]{F1F1F1} 0.103 & {\cellcolor[HTML]{B40426}} \color[HTML]{F1F1F1} 0.178 & {\cellcolor[HTML]{3B4CC0}} \color[HTML]{F1F1F1} -0.009 & {\cellcolor[HTML]{B40426}} \color[HTML]{F1F1F1} 0.067 & {\cellcolor[HTML]{3B4CC0}} \color[HTML]{F1F1F1} -0.005 & {\cellcolor[HTML]{DC5D4A}} \color[HTML]{F1F1F1} 0.011 \\
Boosting & {\cellcolor[HTML]{3B4CC0}} \color[HTML]{F1F1F1} 0.024 & {\cellcolor[HTML]{3B4CC0}} \color[HTML]{F1F1F1} 0.013 & {\cellcolor[HTML]{9BBCFF}} \color[HTML]{000000} -0.000 & {\cellcolor[HTML]{B6CEFA}} \color[HTML]{000000} 0.036 & {\cellcolor[HTML]{799CF8}} \color[HTML]{F1F1F1} 0.002 & {\cellcolor[HTML]{D5DBE5}} \color[HTML]{000000} 0.004 \\
MLP & {\cellcolor[HTML]{6687ED}} \color[HTML]{F1F1F1} 0.035 & {\cellcolor[HTML]{4257C9}} \color[HTML]{F1F1F1} 0.017 & {\cellcolor[HTML]{B1CBFC}} \color[HTML]{000000} 0.002 & {\cellcolor[HTML]{3B4CC0}} \color[HTML]{F1F1F1} 0.018 & {\cellcolor[HTML]{D1493F}} \color[HTML]{F1F1F1} 0.028 & {\cellcolor[HTML]{B40426}} \color[HTML]{F1F1F1} 0.013 \\
Trans & {\cellcolor[HTML]{E1DAD6}} \color[HTML]{000000} 0.065 & {\cellcolor[HTML]{ABC8FD}} \color[HTML]{000000} 0.069 & {\cellcolor[HTML]{B40426}} \color[HTML]{F1F1F1} 0.022 & {\cellcolor[HTML]{84A7FC}} \color[HTML]{F1F1F1} 0.029 & {\cellcolor[HTML]{B40426}} \color[HTML]{F1F1F1} 0.031 & {\cellcolor[HTML]{3B4CC0}} \color[HTML]{F1F1F1} -0.004 \\
\bottomrule
\end{tabular}
\vskip 0.05in
\caption{Model performance on superposition of effects, $\rho = 0.05$}
    \label{tab:model_performance1}
    \end{table*}

\begin{table*}[t]
    \centering
    \begin{tabular}{lcccccc}
\toprule
Effect & Optimal & Linear & Fea-Nonlin & TS-Shift & CS-Shift & TSCS-Shift \\
\midrule
Lasso & {\cellcolor[HTML]{B40426}} \color[HTML]{F1F1F1} 0.605 & {\cellcolor[HTML]{B40426}} \color[HTML]{F1F1F1} 0.590 & {\cellcolor[HTML]{516DDB}} \color[HTML]{F1F1F1} 0.021 & {\cellcolor[HTML]{B40426}} \color[HTML]{F1F1F1} 0.594 & {\cellcolor[HTML]{F59D7E}} \color[HTML]{000000} 0.084 & {\cellcolor[HTML]{B40426}} \color[HTML]{F1F1F1} 0.085 \\
Boosting & {\cellcolor[HTML]{8DB0FE}} \color[HTML]{000000} 0.257 & {\cellcolor[HTML]{ABC8FD}} \color[HTML]{000000} 0.251 & {\cellcolor[HTML]{4F69D9}} \color[HTML]{F1F1F1} 0.020 & {\cellcolor[HTML]{ADC9FD}} \color[HTML]{000000} 0.251 & {\cellcolor[HTML]{3B4CC0}} \color[HTML]{F1F1F1} 0.033 & {\cellcolor[HTML]{6F92F3}} \color[HTML]{F1F1F1} 0.029 \\
MLP & {\cellcolor[HTML]{3B4CC0}} \color[HTML]{F1F1F1} 0.139 & {\cellcolor[HTML]{3B4CC0}} \color[HTML]{F1F1F1} 0.078 & {\cellcolor[HTML]{3B4CC0}} \color[HTML]{F1F1F1} 0.013 & {\cellcolor[HTML]{3B4CC0}} \color[HTML]{F1F1F1} 0.072 & {\cellcolor[HTML]{E2DAD5}} \color[HTML]{000000} 0.069 & {\cellcolor[HTML]{B8122A}} \color[HTML]{F1F1F1} 0.084 \\
Trans & {\cellcolor[HTML]{7699F6}} \color[HTML]{F1F1F1} 0.225 & {\cellcolor[HTML]{6E90F2}} \color[HTML]{F1F1F1} 0.160 & {\cellcolor[HTML]{B40426}} \color[HTML]{F1F1F1} 0.118 & {\cellcolor[HTML]{536EDD}} \color[HTML]{F1F1F1} 0.113 & {\cellcolor[HTML]{B40426}} \color[HTML]{F1F1F1} 0.102 & {\cellcolor[HTML]{3B4CC0}} \color[HTML]{F1F1F1} 0.018 \\
\bottomrule
\end{tabular}
\vskip 0.05in
\caption{Model performance on superposition of effects, $\rho = 0.2$}
    \label{tab:model_performance2}
    \end{table*}

\section{Dynamic attention sparsity}\label{sparsity}

\subsection{Motivation}\label{sparsity_motiv}

The experiments presented in the previous section suggest several intuitive properties that an optimal attention matrix should satisfy, depending on the structure of the underlying data-generating process. To make these intuitions explicit and simplify exposition, we focus on the transformer architecture composed of one temporal attention layer and one cross-sectional attention layer, i.e. a \texttt{"TC"} model in the notations of Section \ref{subsec_models}. We further restrict attention to a single head per layer, which allows direct visualization of the learned attention matrices. Importantly, the sparsity mechanisms discussed below naturally extend to deeper architectures and multi-head attention, although we do not explore these extensions here.

Several simple cases illustrate why sparsity may arise as a natural inductive bias for attention:
\begin{itemize}
    \item \textbf{Temporal lag structure.} In the presence of a single predictive temporal lag (e.g., $s_{j,n}=1$ in \eqref{CS_shift}), the optimal temporal attention matrix is sparse, with non-zero mass concentrated on the first sub-diagonal.
    \item \textbf{Multiple temporal lags.} If two lags carry predictive information (e.g., $s_{j,n}=1$ or $2$), the optimal attention matrix exhibits sparsity concentrated on the first two sub-diagonals.
    \item \textbf{Cross-sectional dependence.} When a single time series among many has predictive power for the target, the optimal cross-sectional attention matrix contains exactly one non-zero entry per row, corresponding to that series. The identity of this series is determined by the data-generating process and must be learned from the data.
    \item \textbf{Mixed effects.} In more general settings combining temporal and cross-sectional dependencies, attention matrices need not be sparse. In such cases, enforcing sparsity deterministically would be suboptimal.
\end{itemize}

These examples highlight a central challenge: while sparsity is often desirable, its structure cannot be specified without a prior and may vary substantially across regimes. This motivates the need for a mechanism that can adaptively sparsify attention when warranted by the data, while retaining dense attention when necessary.

\subsection{Max attention algorithm}\label{sparsity_algo}

To improve robustness in low signal-to-noise regimes, we introduce a dynamic sparsification mechanism within the scaled dot-product attention computation. Unlike deterministic sparse attention patterns introduced in prior work (e.g., \cite{child2019generating}), our approach is fully data-driven and does not assume a known sparsity structure. Moreover, it allows the effective sparsity level to vary across attention rows and across learning regimes.

The key idea is to treat each attention row as a discrete probability distribution and to assess the significance of its entries relative to the dominant mass in that row, rather than enforcing a fixed number of non-zero coefficients.

Given query, key, and value tensors \( Q \), \( K \), and \( V \), we compute the attention logits
\[
A = \frac{QK^\top}{\sqrt{d}} + b,
\]
where \( d \) denotes the head dimension and \( b \) is a bias tensor encoding any attention mask (e.g., a causal mask with \( b_{i,j} = -\infty \) for \( j > i \)).

We apply a softmax to compute attention probabilities and average these probabilities across the batch dimension, yielding an average attention matrix \( \bar{P} \). For each row of \( \bar{P} \), corresponding to a fixed head and query position, we compute its maximum entry \( M \). Entries smaller than \( K \cdot M \), where \( 0 < K < 1 \) is a threshold parameter, are deemed insignificant and removed. This produces a binary mask \( m \).

Using this mask, we construct an auxiliary bias tensor \( b' \), initialized to zero and set to \( -\infty \) wherever \( m = 0 \). The masked logits are then given by
\[
A' = A + b',
\]
with broadcasting over the batch dimension. The final attention weights are obtained by applying a softmax to \( A' \), and are subsequently used to compute the weighted sum of values \( V \).

In all experiments, we set \( K = 0.1 \). This parameter could also be learned during training or selected via cross-validation. We do not perform this optimization here.

Several alternative sparsification strategies—such as retaining the top-\( k \) entries per row or thresholding based on quantiles—were considered. We found these approaches less suitable for two reasons. First, they impose a fixed sparsity level, implicitly assuming prior knowledge of the true dependency structure. Second, under causal masking, the effective length of attention rows varies with the query position, leading such methods to remain dense near the top of the attention matrix while becoming sparse closer to the bottom.

Our design is motivated by simple limiting cases. In a pure lag-1 TS-shift, the dominant sub-diagonal entry converges toward probability one, while all others remain close to zero. Introducing a second predictive lag leads the two corresponding entries to converge toward equal mass (e.g., 0.5 each). These cases illustrate the need for a relative, rather than absolute, sparsification criterion.

Overall, this procedure adaptively filters weak or noisy attention links while preserving high-confidence dependencies, without imposing a fixed sparsity pattern.

\subsection{Bootstrap results}\label{sparsity_results}

To isolate the effect of dynamic sparsification, we first consider the simplest setting of a lag-1 TS-Shift. As discussed above, the optimal temporal attention matrix in this case contains a single non-zero sub-diagonal. We compare three otherwise identical transformer configurations:
\begin{itemize}
    \item A full attention \textbf{full\_attention} transformer with no enforced sparsity.
    \item A deterministically sparse \textbf{deterministic\_sparse} transformer in which the temporal attention matrix is fixed to its optimal structure. Since this requires knowledge of the data-generating process, it serves as an upper bound on achievable performance.
    \item A transformer equipped with the dynamic sparsification algorithm \textbf{max\_sparse} described above.
\end{itemize}

To assess statistical significance, we perform a two-sample hypothesis test comparing the distribution of out-of-sample correlations obtained by the full-attention model and by the dynamically sparse model. Each sample corresponds to a single iteration of a bootstrap in which the dataset is fully regenerated with identical statistical properties. The significance of the means of the populations is assessed using the p-value of the test \textbf{p\_value}. We repeat this procedure across a range of signal-to-noise ratios in order to quantify how the benefits of sparsity depend on the difficulty of the forecasting problem. For each bootstrapped performance metric we consider a number of samples $n = 90$.

We start by considering the case of the TS-Shift effect with lag 1 for which we have an upper bound benchmark of the model with a deterministic sparsity on the sub-diagonal in Table \ref{tab:shift_ts_sparse}.

\begin{table}[t]
    \centering
\begin{tabular}{lccc}
\toprule
$\rho$ & 0.015 & 0.03 & 0.1 \\
\midrule
full\_attention & {\cellcolor[HTML]{3B4CC0}} \color[HTML]{F1F1F1} 0.017 & {\cellcolor[HTML]{3B4CC0}} \color[HTML]{F1F1F1} 0.071 & {\cellcolor[HTML]{3B4CC0}} \color[HTML]{F1F1F1} 0.455 \\
max\_sparse & {\cellcolor[HTML]{5B7AE5}} \color[HTML]{F1F1F1} 0.025 & {\cellcolor[HTML]{5673E0}} \color[HTML]{F1F1F1} 0.083 & {\cellcolor[HTML]{516DDB}} \color[HTML]{F1F1F1} 0.456 \\
deterministic\_sparse & {\cellcolor[HTML]{B40426}} \color[HTML]{F1F1F1} 0.096 & {\cellcolor[HTML]{B40426}} \color[HTML]{F1F1F1} 0.200 & {\cellcolor[HTML]{B40426}} \color[HTML]{F1F1F1} 0.469 \\
p\_value & 0.028 & 0.126 & 0.467 \\
\bottomrule
\end{tabular}
\vskip 0.05in
\caption{Sparsity performance for effect TS-Shift}
    \label{tab:shift_ts_sparse}
\end{table}

As expected, a deterministic sparsity informed by the ground truth generating process gives the best results. It can be noted that the p-value computed to consider the difference of performance between \textbf{max\_sparse} and \textbf{full\_attention} shows high level of significance for lower correlations and monotonically increases with $\rho$.

We can then consider a similar setup, the CS-Shift effect for which we don't have a readily informed deterministic sparse structure given the ground truth model in Table \ref{tab:shift_cs_sparse}. We see very similar results with a $10\%$ significance for both $\rho = 0.015$ and $\rho = 0.03$ with the same monotonic increase in p-value.

\begin{table}[t]
    \centering
\begin{tabular}{lccc}
\toprule
$\rho$ & 0.015 & 0.03 & 0.1 \\
\midrule
full\_attention & {\cellcolor[HTML]{3B4CC0}} \color[HTML]{F1F1F1} 0.012 & {\cellcolor[HTML]{3B4CC0}} \color[HTML]{F1F1F1} 0.034 & {\cellcolor[HTML]{3B4CC0}} \color[HTML]{F1F1F1} 0.101 \\
max\_sparse & {\cellcolor[HTML]{B40426}} \color[HTML]{F1F1F1} 0.019 & {\cellcolor[HTML]{B40426}} \color[HTML]{F1F1F1} 0.038 & {\cellcolor[HTML]{B40426}} \color[HTML]{F1F1F1} 0.102 \\
p\_value & 0.021 & 0.094 & 0.463 \\
\bottomrule
\end{tabular}
\vskip 0.05in
\caption{Sparsity performance for effect CS-Shift}
    \label{tab:shift_cs_sparse}
\end{table}

Now considering both the superposition of effects and the simple linear effects, we don't expect much added value from a learned sparisty but one can hope that the model remains competitive with a performance that can be compared with a dense attention matrix. This is indeed what we observe as highlighted in Table \ref{tab:all_sparse} and Table \ref{tab:lin_sparse} which shows the robustness and the ability to learn the sparsity structure of a given prediction problem.

\begin{table}[t]
    \centering
\begin{tabular}{lccc}
\toprule
$\rho$ & 0.015 & 0.03 & 0.1 \\
\midrule
full\_attention & {\cellcolor[HTML]{B40426}} \color[HTML]{F1F1F1} 0.035 & {\cellcolor[HTML]{3B4CC0}} \color[HTML]{F1F1F1} 0.075 & {\cellcolor[HTML]{B40426}} \color[HTML]{F1F1F1} 0.243 \\
max\_sparse & {\cellcolor[HTML]{3B4CC0}} \color[HTML]{F1F1F1} 0.034 & {\cellcolor[HTML]{B40426}} \color[HTML]{F1F1F1} 0.075 & {\cellcolor[HTML]{3B4CC0}} \color[HTML]{F1F1F1} 0.231 \\
p\_value & 0.638 & 0.464 & 0.953 \\
\bottomrule
\end{tabular}
\vskip 0.05in
\caption{Sparsity performance on superposition of effects}
\label{tab:all_sparse}
\end{table}

\begin{table}[t]
    \centering
\begin{tabular}{lccc}
\toprule
$\rho$ & 0.015 & 0.03 & 0.1 \\
\midrule
full\_attention & {\cellcolor[HTML]{B40426}} \color[HTML]{F1F1F1} 0.135 & {\cellcolor[HTML]{B40426}} \color[HTML]{F1F1F1} 0.255 & {\cellcolor[HTML]{B40426}} \color[HTML]{F1F1F1} 0.538 \\
max\_sparse & {\cellcolor[HTML]{3B4CC0}} \color[HTML]{F1F1F1} 0.120 & {\cellcolor[HTML]{3B4CC0}} \color[HTML]{F1F1F1} 0.245 & {\cellcolor[HTML]{3B4CC0}} \color[HTML]{F1F1F1} 0.530 \\
p\_value & 0.902 & 0.779 & 0.724 \\
\bottomrule
\end{tabular}
\vskip 0.05in
\caption{Sparsity performance for Linear effect}
    \label{tab:lin_sparse}
\end{table}

As a way to visualize what happens in the temporal and cross-sectional cases, we present in Appendix \ref{app_attention} the attention matrix learned within each generating process for one random example of the bootstrap samples. The temporal attention matrix for a TS-Shift effect is shown in Table \ref{tab:attn_ts} and the cross-sectional attention matrix for a CS-Shift effect is shown in Table \ref{tab:attn_cs}. Note that while the temporal attention shape is fairly intuitive, the cross-sectional attention would have only one non-zero coefficient per row in the absence of noise. We see here how the sparsity helps to remove a part of the noise for both examples (often one clear dominating coefficient per row in the CS case and the sub-diagonal of the temporal matrix with high probability weight).

\section{Conclusion and outlooks}\label{sec_conclu}

In this work, we adopted a statistical approach based on synthetically generated canonical effects to study both the performance and the underlying mechanisms of transformer architectures for multivariate time-series forecasting. Our objective was to help bridge the gap between the rapidly growing empirical literature on transformers—often centered on fixed benchmark real-world datasets—and more interpretable statistical analysis that facilitates mechanistic understanding and generalization to complex settings.
Our results show that, even in regimes characterized by extremely low signal-to-noise ratios, transformer architectures with stacked temporal and cross-sectional attention layers can outperform traditional time-series models. At the same time, we observe that interactions between these two dimensions — corresponding in practice to shifts simultaneously in the time-series and cross-sectional directions — remain more challenging to capture, even when multiple attention layers are combined. This limitation points to promising directions for future research on how different attention dimensions should be composed or integrated.
Motivated by an analogy with the curse of dimensionality in classical statistics, we introduced a dynamic sparsification mechanism that can be applied independently to each attention layer and head. Empirically, this mechanism yields significant improvements over dense attention in settings where sparsity is optimal, while remaining competitive when the data-generating process does not exhibit sparse structure. These benefits are particularly pronounced in very low signal-to-noise regimes.
Beyond the specific architectural insights, our study highlights the value of synthetic data and controlled experimental designs for assessing statistical significance and inductive biases in modern deep learning models, especially in settings with limited data. We believe this perspective may inform both real-world time-series forecasting applications and future methodological work aimed at understanding and improving transformer-based models.


\vspace{0.5cm}
GitHub repository available at:
\newline
\url{https://github.com/cyrilgarcia009/TSNN}

\bibliographystyle{icml2026}
\bibliography{references}

\appendix
\onecolumn

\section{Examples of sparse attention matrices}\label{app_attention}

\begin{table*}[h]
    \centering
\begin{tabular}{lcccccccccc}
\toprule
 & 0 & 1 & 2 & 3 & 4 & 5 & 6 & 7 & 8 & 9 \\
\midrule
0 & {\cellcolor[HTML]{023858}} \color[HTML]{F1F1F1} 1.000 & {\cellcolor[HTML]{FFF7FB}} \color[HTML]{000000} 0.000 & {\cellcolor[HTML]{FFF7FB}} \color[HTML]{000000} 0.000 & {\cellcolor[HTML]{FFF7FB}} \color[HTML]{000000} 0.000 & {\cellcolor[HTML]{FFF7FB}} \color[HTML]{000000} 0.000 & {\cellcolor[HTML]{FFF7FB}} \color[HTML]{000000} 0.000 & {\cellcolor[HTML]{FFF7FB}} \color[HTML]{000000} 0.000 & {\cellcolor[HTML]{FFF7FB}} \color[HTML]{000000} 0.000 & {\cellcolor[HTML]{FFF7FB}} \color[HTML]{000000} 0.000 & {\cellcolor[HTML]{FFF7FB}} \color[HTML]{000000} 0.000 \\
1 & {\cellcolor[HTML]{2C89BD}} \color[HTML]{F1F1F1} 0.650 & {\cellcolor[HTML]{529BC7}} \color[HTML]{F1F1F1} 0.350 & {\cellcolor[HTML]{FFF7FB}} \color[HTML]{000000} 0.000 & {\cellcolor[HTML]{FFF7FB}} \color[HTML]{000000} 0.000 & {\cellcolor[HTML]{FFF7FB}} \color[HTML]{000000} 0.000 & {\cellcolor[HTML]{FFF7FB}} \color[HTML]{000000} 0.000 & {\cellcolor[HTML]{FFF7FB}} \color[HTML]{000000} 0.000 & {\cellcolor[HTML]{FFF7FB}} \color[HTML]{000000} 0.000 & {\cellcolor[HTML]{FFF7FB}} \color[HTML]{000000} 0.000 & {\cellcolor[HTML]{FFF7FB}} \color[HTML]{000000} 0.000 \\
2 & {\cellcolor[HTML]{E8E4F0}} \color[HTML]{000000} 0.141 & {\cellcolor[HTML]{023858}} \color[HTML]{F1F1F1} 0.616 & {\cellcolor[HTML]{A4BCDA}} \color[HTML]{000000} 0.243 & {\cellcolor[HTML]{FFF7FB}} \color[HTML]{000000} 0.000 & {\cellcolor[HTML]{FFF7FB}} \color[HTML]{000000} 0.000 & {\cellcolor[HTML]{FFF7FB}} \color[HTML]{000000} 0.000 & {\cellcolor[HTML]{FFF7FB}} \color[HTML]{000000} 0.000 & {\cellcolor[HTML]{FFF7FB}} \color[HTML]{000000} 0.000 & {\cellcolor[HTML]{FFF7FB}} \color[HTML]{000000} 0.000 & {\cellcolor[HTML]{FFF7FB}} \color[HTML]{000000} 0.000 \\
3 & {\cellcolor[HTML]{FFF7FB}} \color[HTML]{000000} 0.000 & {\cellcolor[HTML]{D9D8EA}} \color[HTML]{000000} 0.129 & {\cellcolor[HTML]{023858}} \color[HTML]{F1F1F1} 0.637 & {\cellcolor[HTML]{BCC7E1}} \color[HTML]{000000} 0.235 & {\cellcolor[HTML]{FFF7FB}} \color[HTML]{000000} 0.000 & {\cellcolor[HTML]{FFF7FB}} \color[HTML]{000000} 0.000 & {\cellcolor[HTML]{FFF7FB}} \color[HTML]{000000} 0.000 & {\cellcolor[HTML]{FFF7FB}} \color[HTML]{000000} 0.000 & {\cellcolor[HTML]{FFF7FB}} \color[HTML]{000000} 0.000 & {\cellcolor[HTML]{FFF7FB}} \color[HTML]{000000} 0.000 \\
4 & {\cellcolor[HTML]{FFF7FB}} \color[HTML]{000000} 0.000 & {\cellcolor[HTML]{FFF7FB}} \color[HTML]{000000} 0.000 & {\cellcolor[HTML]{FFF7FB}} \color[HTML]{000000} 0.000 & {\cellcolor[HTML]{023858}} \color[HTML]{F1F1F1} 0.751 & {\cellcolor[HTML]{8CB3D5}} \color[HTML]{000000} 0.249 & {\cellcolor[HTML]{FFF7FB}} \color[HTML]{000000} 0.000 & {\cellcolor[HTML]{FFF7FB}} \color[HTML]{000000} 0.000 & {\cellcolor[HTML]{FFF7FB}} \color[HTML]{000000} 0.000 & {\cellcolor[HTML]{FFF7FB}} \color[HTML]{000000} 0.000 & {\cellcolor[HTML]{FFF7FB}} \color[HTML]{000000} 0.000 \\
5 & {\cellcolor[HTML]{FFF7FB}} \color[HTML]{000000} 0.000 & {\cellcolor[HTML]{FFF7FB}} \color[HTML]{000000} 0.000 & {\cellcolor[HTML]{F1EBF4}} \color[HTML]{000000} 0.061 & {\cellcolor[HTML]{E0DEED}} \color[HTML]{000000} 0.135 & {\cellcolor[HTML]{023858}} \color[HTML]{F1F1F1} 0.567 & {\cellcolor[HTML]{AFC1DD}} \color[HTML]{000000} 0.237 & {\cellcolor[HTML]{FFF7FB}} \color[HTML]{000000} 0.000 & {\cellcolor[HTML]{FFF7FB}} \color[HTML]{000000} 0.000 & {\cellcolor[HTML]{FFF7FB}} \color[HTML]{000000} 0.000 & {\cellcolor[HTML]{FFF7FB}} \color[HTML]{000000} 0.000 \\
6 & {\cellcolor[HTML]{FFF7FB}} \color[HTML]{000000} 0.000 & {\cellcolor[HTML]{FFF7FB}} \color[HTML]{000000} 0.000 & {\cellcolor[HTML]{FFF7FB}} \color[HTML]{000000} 0.000 & {\cellcolor[HTML]{FFF7FB}} \color[HTML]{000000} 0.000 & {\cellcolor[HTML]{DCDAEB}} \color[HTML]{000000} 0.112 & {\cellcolor[HTML]{023858}} \color[HTML]{F1F1F1} 0.682 & {\cellcolor[HTML]{BBC7E0}} \color[HTML]{000000} 0.206 & {\cellcolor[HTML]{FFF7FB}} \color[HTML]{000000} 0.000 & {\cellcolor[HTML]{FFF7FB}} \color[HTML]{000000} 0.000 & {\cellcolor[HTML]{FFF7FB}} \color[HTML]{000000} 0.000 \\
7 & {\cellcolor[HTML]{FFF7FB}} \color[HTML]{000000} 0.000 & {\cellcolor[HTML]{FFF7FB}} \color[HTML]{000000} 0.000 & {\cellcolor[HTML]{FFF7FB}} \color[HTML]{000000} 0.000 & {\cellcolor[HTML]{FFF7FB}} \color[HTML]{000000} 0.000 & {\cellcolor[HTML]{FFF7FB}} \color[HTML]{000000} 0.000 & {\cellcolor[HTML]{DDDBEC}} \color[HTML]{000000} 0.133 & {\cellcolor[HTML]{023858}} \color[HTML]{F1F1F1} 0.658 & {\cellcolor[HTML]{B4C4DF}} \color[HTML]{000000} 0.210 & {\cellcolor[HTML]{FFF7FB}} \color[HTML]{000000} 0.000 & {\cellcolor[HTML]{FFF7FB}} \color[HTML]{000000} 0.000 \\
8 & {\cellcolor[HTML]{FFF7FB}} \color[HTML]{000000} 0.000 & {\cellcolor[HTML]{FFF7FB}} \color[HTML]{000000} 0.000 & {\cellcolor[HTML]{FFF7FB}} \color[HTML]{000000} 0.000 & {\cellcolor[HTML]{FFF7FB}} \color[HTML]{000000} 0.000 & {\cellcolor[HTML]{FFF7FB}} \color[HTML]{000000} 0.000 & {\cellcolor[HTML]{FFF7FB}} \color[HTML]{000000} 0.000 & {\cellcolor[HTML]{D8D7E9}} \color[HTML]{000000} 0.141 & {\cellcolor[HTML]{023858}} \color[HTML]{F1F1F1} 0.631 & {\cellcolor[HTML]{9EBAD9}} \color[HTML]{000000} 0.228 & {\cellcolor[HTML]{FFF7FB}} \color[HTML]{000000} 0.000 \\
9 & {\cellcolor[HTML]{FFF7FB}} \color[HTML]{000000} 0.000 & {\cellcolor[HTML]{FFF7FB}} \color[HTML]{000000} 0.000 & {\cellcolor[HTML]{FFF7FB}} \color[HTML]{000000} 0.000 & {\cellcolor[HTML]{FFF7FB}} \color[HTML]{000000} 0.000 & {\cellcolor[HTML]{FFF7FB}} \color[HTML]{000000} 0.000 & {\cellcolor[HTML]{FFF7FB}} \color[HTML]{000000} 0.000 & {\cellcolor[HTML]{EEE8F3}} \color[HTML]{000000} 0.075 & {\cellcolor[HTML]{D5D5E8}} \color[HTML]{000000} 0.144 & {\cellcolor[HTML]{023858}} \color[HTML]{F1F1F1} 0.573 & {\cellcolor[HTML]{023858}} \color[HTML]{F1F1F1} 0.209 \\
\bottomrule
\end{tabular}
\vskip 0.05in
\caption{Temporal learned attention matrix for $\rho = 0.03$}
    \label{tab:attn_ts}
\end{table*}

\begin{table*}[h]
    \centering
\begin{tabular}{lcccccccccc}
\toprule
 & 0 & 1 & 2 & 3 & 4 & 5 & 6 & 7 & 8 & 9 \\
\midrule
0 & {\cellcolor[HTML]{FFF7FB}} \color[HTML]{000000} 0.000 & {\cellcolor[HTML]{FFF7FB}} \color[HTML]{000000} 0.000 & {\cellcolor[HTML]{FFF7FB}} \color[HTML]{000000} 0.000 & {\cellcolor[HTML]{FFF7FB}} \color[HTML]{000000} 0.000 & {\cellcolor[HTML]{FFF7FB}} \color[HTML]{000000} 0.000 & {\cellcolor[HTML]{FFF7FB}} \color[HTML]{000000} 0.000 & {\cellcolor[HTML]{FFF7FB}} \color[HTML]{000000} 0.000 & {\cellcolor[HTML]{FFF7FB}} \color[HTML]{000000} 0.000 & {\cellcolor[HTML]{023858}} \color[HTML]{F1F1F1} 0.714 & {\cellcolor[HTML]{9CB9D9}} \color[HTML]{000000} 0.286 \\
1 & {\cellcolor[HTML]{FFF7FB}} \color[HTML]{000000} 0.000 & {\cellcolor[HTML]{FFF7FB}} \color[HTML]{000000} 0.000 & {\cellcolor[HTML]{FFF7FB}} \color[HTML]{000000} 0.000 & {\cellcolor[HTML]{C0C9E2}} \color[HTML]{000000} 0.135 & {\cellcolor[HTML]{023858}} \color[HTML]{F1F1F1} 0.453 & {\cellcolor[HTML]{BCC7E1}} \color[HTML]{000000} 0.140 & {\cellcolor[HTML]{FFF7FB}} \color[HTML]{000000} 0.000 & {\cellcolor[HTML]{FFF7FB}} \color[HTML]{000000} 0.000 & {\cellcolor[HTML]{4094C3}} \color[HTML]{F1F1F1} 0.273 & {\cellcolor[HTML]{FFF7FB}} \color[HTML]{000000} 0.000 \\
2 & {\cellcolor[HTML]{2F8BBE}} \color[HTML]{F1F1F1} 0.245 & {\cellcolor[HTML]{FFF7FB}} \color[HTML]{000000} 0.000 & {\cellcolor[HTML]{FFF7FB}} \color[HTML]{000000} 0.000 & {\cellcolor[HTML]{FFF7FB}} \color[HTML]{000000} 0.000 & {\cellcolor[HTML]{FFF7FB}} \color[HTML]{000000} 0.000 & {\cellcolor[HTML]{96B6D7}} \color[HTML]{000000} 0.157 & {\cellcolor[HTML]{FFF7FB}} \color[HTML]{000000} 0.000 & {\cellcolor[HTML]{4E9AC6}} \color[HTML]{F1F1F1} 0.218 & {\cellcolor[HTML]{FFF7FB}} \color[HTML]{000000} 0.000 & {\cellcolor[HTML]{023858}} \color[HTML]{F1F1F1} 0.380 \\
3 & {\cellcolor[HTML]{ECE7F2}} \color[HTML]{000000} 0.030 & {\cellcolor[HTML]{023858}} \color[HTML]{F1F1F1} 0.242 & {\cellcolor[HTML]{78ABD0}} \color[HTML]{F1F1F1} 0.118 & {\cellcolor[HTML]{DCDAEB}} \color[HTML]{000000} 0.048 & {\cellcolor[HTML]{FFF7FB}} \color[HTML]{000000} 0.000 & {\cellcolor[HTML]{045585}} \color[HTML]{F1F1F1} 0.216 & {\cellcolor[HTML]{FFF7FB}} \color[HTML]{000000} 0.000 & {\cellcolor[HTML]{0D75B3}} \color[HTML]{F1F1F1} 0.176 & {\cellcolor[HTML]{ECE7F2}} \color[HTML]{000000} 0.030 & {\cellcolor[HTML]{4E9AC6}} \color[HTML]{F1F1F1} 0.139 \\
4 & {\cellcolor[HTML]{023858}} \color[HTML]{F1F1F1} 0.526 & {\cellcolor[HTML]{FFF7FB}} \color[HTML]{000000} 0.000 & {\cellcolor[HTML]{FFF7FB}} \color[HTML]{000000} 0.000 & {\cellcolor[HTML]{FFF7FB}} \color[HTML]{000000} 0.000 & {\cellcolor[HTML]{FFF7FB}} \color[HTML]{000000} 0.000 & {\cellcolor[HTML]{79ABD0}} \color[HTML]{F1F1F1} 0.256 & {\cellcolor[HTML]{CCCFE5}} \color[HTML]{000000} 0.139 & {\cellcolor[HTML]{FFF7FB}} \color[HTML]{000000} 0.000 & {\cellcolor[HTML]{FFF7FB}} \color[HTML]{000000} 0.000 & {\cellcolor[HTML]{E7E3F0}} \color[HTML]{000000} 0.079 \\
5 & {\cellcolor[HTML]{DCDAEB}} \color[HTML]{000000} 0.047 & {\cellcolor[HTML]{023858}} \color[HTML]{F1F1F1} 0.237 & {\cellcolor[HTML]{A8BEDC}} \color[HTML]{000000} 0.087 & {\cellcolor[HTML]{056FAF}} \color[HTML]{F1F1F1} 0.179 & {\cellcolor[HTML]{DDDBEC}} \color[HTML]{000000} 0.046 & {\cellcolor[HTML]{C4CBE3}} \color[HTML]{000000} 0.068 & {\cellcolor[HTML]{9AB8D8}} \color[HTML]{000000} 0.096 & {\cellcolor[HTML]{FFF7FB}} \color[HTML]{000000} 0.000 & {\cellcolor[HTML]{4496C3}} \color[HTML]{F1F1F1} 0.141 & {\cellcolor[HTML]{96B6D7}} \color[HTML]{000000} 0.098 \\
6 & {\cellcolor[HTML]{023858}} \color[HTML]{F1F1F1} 0.272 & {\cellcolor[HTML]{FFF7FB}} \color[HTML]{000000} 0.000 & {\cellcolor[HTML]{529BC7}} \color[HTML]{F1F1F1} 0.155 & {\cellcolor[HTML]{62A2CB}} \color[HTML]{F1F1F1} 0.146 & {\cellcolor[HTML]{FFF7FB}} \color[HTML]{000000} 0.000 & {\cellcolor[HTML]{E0DEED}} \color[HTML]{000000} 0.049 & {\cellcolor[HTML]{97B7D7}} \color[HTML]{000000} 0.112 & {\cellcolor[HTML]{FFF7FB}} \color[HTML]{000000} 0.000 & {\cellcolor[HTML]{023C5F}} \color[HTML]{F1F1F1} 0.267 & {\cellcolor[HTML]{FFF7FB}} \color[HTML]{000000} 0.000 \\
7 & {\cellcolor[HTML]{E5E1EF}} \color[HTML]{000000} 0.081 & {\cellcolor[HTML]{FFF7FB}} \color[HTML]{000000} 0.000 & {\cellcolor[HTML]{EAE6F1}} \color[HTML]{000000} 0.069 & {\cellcolor[HTML]{FFF7FB}} \color[HTML]{000000} 0.000 & {\cellcolor[HTML]{5A9EC9}} \color[HTML]{F1F1F1} 0.279 & {\cellcolor[HTML]{023858}} \color[HTML]{F1F1F1} 0.506 & {\cellcolor[HTML]{FFF7FB}} \color[HTML]{000000} 0.000 & {\cellcolor[HTML]{FFF7FB}} \color[HTML]{000000} 0.000 & {\cellcolor[HTML]{ECE7F2}} \color[HTML]{000000} 0.065 & {\cellcolor[HTML]{FFF7FB}} \color[HTML]{000000} 0.000 \\
8 & {\cellcolor[HTML]{8FB4D6}} \color[HTML]{000000} 0.116 & {\cellcolor[HTML]{023858}} \color[HTML]{F1F1F1} 0.269 & {\cellcolor[HTML]{FFF7FB}} \color[HTML]{000000} 0.000 & {\cellcolor[HTML]{0A73B2}} \color[HTML]{F1F1F1} 0.198 & {\cellcolor[HTML]{A9BFDC}} \color[HTML]{000000} 0.098 & {\cellcolor[HTML]{EAE6F1}} \color[HTML]{000000} 0.036 & {\cellcolor[HTML]{D9D8EA}} \color[HTML]{000000} 0.056 & {\cellcolor[HTML]{D7D6E9}} \color[HTML]{000000} 0.059 & {\cellcolor[HTML]{3790C0}} \color[HTML]{F1F1F1} 0.168 & {\cellcolor[HTML]{FFF7FB}} \color[HTML]{000000} 0.000 \\
9 & {\cellcolor[HTML]{C8CDE4}} \color[HTML]{000000} 0.122 & {\cellcolor[HTML]{FFF7FB}} \color[HTML]{000000} 0.000 & {\cellcolor[HTML]{FFF7FB}} \color[HTML]{000000} 0.000 & {\cellcolor[HTML]{FFF7FB}} \color[HTML]{000000} 0.000 & {\cellcolor[HTML]{E8E4F0}} \color[HTML]{000000} 0.064 & {\cellcolor[HTML]{B3C3DE}} \color[HTML]{000000} 0.150 & {\cellcolor[HTML]{023858}} \color[HTML]{F1F1F1} 0.445 & {\cellcolor[HTML]{EFE9F3}} \color[HTML]{000000} 0.048 & {\cellcolor[HTML]{FFF7FB}} \color[HTML]{000000} 0.000 & {\cellcolor[HTML]{A4BCDA}} \color[HTML]{000000} 0.170 \\
\bottomrule
\end{tabular}
\vskip 0.05in
\caption{Cross-sectional learned attention matrix for $\rho = 0.03$}
    \label{tab:attn_cs}
\end{table*}

\section{Analytical derivation of the expected OLS performance}\label{app_theo}

To get a sense of the level of correlation between our model prediction $\widehat Y$ and the optimal prediction $\widetilde Y$ we expect to find as a function of the size of the data (total time steps, number of series and features, number of time-series lags used), we perform an analytic computation in the linear case where there is no time-series structure (i.e. flatten all features).

Let $X_1, \dots, X_N$ be $N$ i.i.d. $\mathcal{N}(0,1)$ random variables. Consider $\widetilde Y := \sum_{i=1}^N \rho_i X_i$ and $\rho^2 := \sum_{i=1}^N \rho^2_i < 1$. Define $Y = \widetilde Y + \sqrt{1 - \rho^2} Z$ with $Z \sim \mathcal{N}(0,1)$ independent of all the $X_i$. Consider now $T$ i.i.d. samples of $Y, X_1, \dots, X_N$ and the multidimensional regression problem:
\begin{align*}
Y_t = \sum_i \beta_i X_{t, i} + \epsilon.
\end{align*}
We can then estimate the $\beta_i$ using OLS as
$\widehat \beta = (X^T X)^{-1} X^T Y$
and from here we obtain the model prediction $\widehat Y = X \widehat \beta$. The goal is thus to understand the distribution of $\text{Correl}(\widehat Y, \widetilde Y )$ as a function of $N,T$. Here we will just compute the expectation.

\vspace{\baselineskip}
\noindent
\textbf{i) In-sample correlation.} We assume that $N \leq T$ so that $\widehat \beta$ is almost surely well-defined. Using $\widehat Y = X \widehat \beta$, $\widetilde Y = X \rho$, we compute:
\begin{align*}
\widehat Y = X (X^T X)^{-1} X^T \widetilde Y + \sqrt{1 - \rho^2} X (X^T X)^{-1} X^T Z  = \widetilde Y + \sqrt{1 - \rho^2} X (X^T X)^{-1} X^T Z.
\end{align*}
Note that $\mathbb{E}[\widetilde Y^2] = T \rho^2$. Then using the independence of $X$ and $Z$:
\begin{align*}
\mathbb{E}[\widetilde Y ^T \widehat Y] &= \mathbb{E}[\rho^{T} X^T X \rho] = T \rho^2,
\end{align*}
and:
\begin{align*}
\mathbb{E}[ \widehat Y ^2] &= T \rho^2 + (1 - \rho^2)  \mathbb{E}[Z^T X (X^T X)^{-1} X^T Z] = T \rho^2 + (1 - \rho^2)  \mathbb{E}[Tr(X (X^T X)^{-1} X^T)].
\end{align*}
The trace in the expectation above simplifies to the trace of the identity matrix of size $N$, which simply gives $N$. Putting everything together we obtain:
\begin{align}
\frac{\mathbb{E}[\widetilde Y ^T \widehat Y]}{\sqrt{\mathbb{E}[ \widetilde Y ^2]  \mathbb{E}[ \widehat Y ^2]}} = \frac{\rho}{\sqrt{\rho^2 + (1- \rho^2) \frac{N}{T}}}.
\end{align}

\vspace{\baselineskip}
\noindent
\textbf{ii) Out-of-sample correlation.} We repeat the above computation but computing this time an out-of-sample correlation. More precisely, consider now $X_o$ a $T_o \times N$ matrix with $\mathcal{N}(0,1)$ i.i.d. entries independent of everything. Setting $\widehat Y_o = X_o \widehat \beta$, $\widetilde Y_o = X_o \rho$, the goal is now to compute $\text{Correl}(\widehat Y_o, \widetilde Y_o)$ as a function of $N, T, T_o$. In this case we get $\mathbb{E}[\widetilde Y_o^2] = T_o \rho^2$, $\mathbb{E}[\widetilde Y_o ^T \widehat Y_o] = T_o \rho^2$, and:
\begin{align*}
\mathbb{E}[ \widehat Y_o ^2] &= T_o \rho^2 + (1 - \rho^2)  \mathbb{E}[Z^T X (X^T X)^{-1} X^T_o X_o (X^T X)^{-1} X^T Z].
\end{align*}
In the last term the expectation is over the randomness of $Z, X, X_o$ which are all independent. The expectation over $Z$ gives a trace as before and we arrive at:
\begin{align*}
\mathbb{E}[ \widehat Y_o ^2] &= T_o \rho^2 + (1 - \rho^2)  \mathbb{E}[Tr(X^T_o X_o (X^T X)^{-1})] = T_o \rho^2 + T_o (1 - \rho^2)  \mathbb{E}[Tr((X^T X)^{-1})].
\end{align*}
To evaluate the last expectation above, we use the Marchenko-Pastur limit, i.e. $N, T \rightarrow \infty$ with $N/T \rightarrow \gamma < 1$, under which $Tr((X^T X)^{-1})$ converges almost surely to $\frac{\gamma}{1 - \gamma}$. Putting everything together we obtain under this limit:
\begin{align}\label{theo_correl_oos}
\frac{\mathbb{E}[\widetilde Y_o ^T \widehat Y_o]}{\sqrt{\mathbb{E}[ \widetilde Y_o ^2]  \mathbb{E}[ \widehat Y_o ^2]}} \rightarrow \frac{\rho}{\sqrt{\rho^2 + (1- \rho^2) \frac{\gamma}{1-\gamma}}}.
\end{align}

\end{document}